# Object Detection and Tracking

**OMAR FARUK**

Bachelor of Engineering in ICT, Comilla University, Bangladesh
Email: S6990161@studenti.unige.it
Omarcou45@gmail.com

## Abstract

Efficient and accurate object detection is an important topic in the development of computer vision systems. With the advent of deep learning techniques, the accuracy of object detection has increased significantly. The project aims to integrate a modern technique for object detection with the aim of achieving high accuracy with real-time performance. The reliance on other computer vision algorithms in many object identification systems, which results in poor and ineffective performance, is a significant obstacle. In this research, we solve the end-to-end object detection problem entirely using deep learning techniques. The network is trained using the most difficult publicly available dataset, which is used for an annual item detection challenge. Applications that need object detection can benefit the system's quick and precise finding.

## 1. Introduction

Object detection is a well-known computer technology related to computer vision and image processing. With the advent of deep learning techniques, the accuracy of object detection has increased significantly. It focuses on detecting objects or instances of a certain class (flowers, animals) in digital images and videos. There are various applications, including face recognition, character recognition, and media calculators.

### 1.1 Problem Statement

A decade ago, many computer vision issues had reached the point of saturation. However, because deep learning techniques have become more popular, these issues' accuracy has considerably increased. The image classifier cation, which is regarded as a predictor of the image's class, is one of the primary issues. Image localization is a rather challenging task where the system must anticipate the location class of a single object in an image (a bounding box around the object). Object detection is the most challenging problem (this project) because it combines both localization and classification. In this instance, a picture will be the input to the system, and the system will produce a bounding box along with the feature that matches every object in the image.

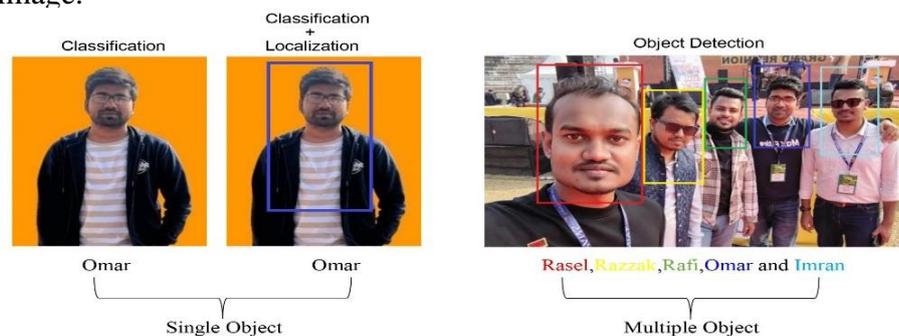

**Fig:1** Object Detection in Image Recognition

*1.2 Applications*

Face recognition is a popular application for object recognition that is used on practically all smartphone cameras. When a significant number of objects need to be recognized for autonomous driving, more widespread (multi-class) applications can be used. It is crucial in surveillance systems as well. These systems can be used in conjunction with various tasks, including B. Estimating the pose. In the first step of the pipeline, it recognizes the object, and in the second stage, it estimates the pose of the area that was recognized. It has uses in robotics and medicine since it can be used to track items. Therefore, there are several applications for this issue. Face recognition is a well-known object identification program that is utilized by nearly all mobile devices.

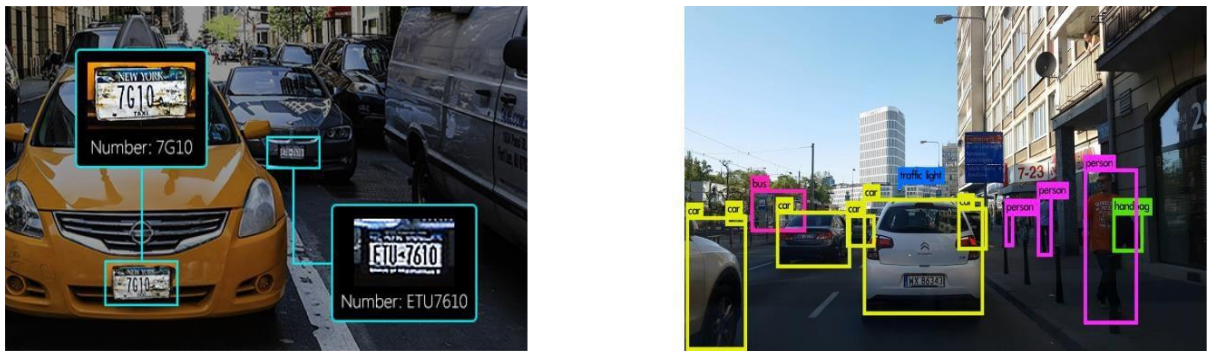

**Fig:2** Vehicle and Object Detection in Traffic

*1.3 Challenges*

The fundamental issue with this problem is that the output's dimensions will change depending on how many objects are or are not present in each input image. The input and output dimensions must be given for the model to be trained for a typical machine learning activity. The requirement for real-time (> 30 fps) while maintaining detection accuracy is a significant barrier to the general use of object detection systems. The longer it takes to guess, the more complicated the model is. The accuracy of the model decreases with decreasing complexity. Depending on the application, a trade-off between accuracy and performance should be made. Regression and classification are involved in the issue, requiring simultaneous model training. This makes the issue more difficult.

**2. Proposed System**

In the realm of computer technology and software systems that traditionally lack the capability to detect or comprehend images and scenes, computer vision emerges as a key area of artificial intelligence. Computer vision encompasses various domains such as image recognition, object detection, image synthesis, and image super-resolution. Among these, object detection holds significant practical value due to its extensive applications in diverse fields.

Object detection refers to a software system's ability to locate and identify individual objects within an image or scene. It has been widely employed in face recognition, vehicle detection, pedestrian counting, web imagery analysis, security systems, and autonomous vehicles. The advancements in deep learning have significantly improved object detection, making it more accurate and reliable for real-world applications.

*2.1 Object Detection Techniques*

Over time, several methodologies have been developed for object detection, each with varying levels of accuracy and efficiency. Traditional approaches relied on feature extraction techniques, such as those provided by OpenCV, a well-known computer vision toolkit. However, these traditional algorithms often struggle to deliver the required performance in complex and dynamic environments.

Modern object detection techniques leverage deep learning frameworks to achieve state-of-the-art performance. Convolutional Neural Networks (CNNs) have revolutionized the field by enabling highly accurate and efficient object detection. Popular architectures include:

I. *Region-Based Convolutional Neural Networks (R-CNN, Fast R-CNN, Faster R-CNN):* These models generate region proposals and classify objects within those regions.
II. *Single Shot MultiBox Detector (SSD):* A real-time object detection framework that balances speed and accuracy.
III. *You Only Look Once (YOLO):* A real-time object detection model that processes entire images in a single pass, making it suitable for real-time applications.
IV. *EfficientDet:* A scalable and optimized object detection model that offers high accuracy with reduced computational overhead.

*2.2 Implementation Details*

This project integrates various libraries, software packages, and deep learning frameworks to implement an efficient object detection system. The following technologies have been utilized:

I. *Programming Language:* Python was selected due to its extensive support for machine learning and deep learning libraries.
II. *Deep Learning Framework:* TensorFlow was used for training and implementing deep learning models. TensorFlow offers both CPU and GPU versions, allowing flexibility in deployment.
III. *Hardware Acceleration:* The project leveraged GPU acceleration to enhance computational efficiency. A GeForce GTX 990 Ti graphics card was employed to optimize deep learning computations. NVIDIA GPUs are widely used for deep learning tasks due to their support for CUDA Toolkit and cuDNN, which significantly accelerate neural network operations.
IV. *Supporting Libraries:* OpenCV for image processing, NumPy for numerical computations, and Matplotlib for visualizing detection outputs.

*2.3 System Workflow*

The proposed object detection system follows a structured workflow to achieve high accuracy and performance. The process consists of the following steps:

1. *Data Collection & Preprocessing:* Gathering datasets, augmenting images, and normalizing pixel values for training.
2. *Model Selection & Training:* Using TensorFlow-based architectures like YOLO or Faster R-CNN to train models on annotated datasets.
3. *Inference & Detection:* Deploying trained models to detect and classify objects in real-

time images and videos.
  4. ***Evaluation & Optimization:*** Assessing performance using metrics such as mean Average Precision (mAP) and optimizing model parameters for enhanced accuracy.

*2.4 Challenges and Future Enhancements*
Despite the effectiveness of modern object detection techniques, several challenges persist:
  I. ***Real-time Performance:*** Ensuring low-latency detection for real-time applications remains a key concern.
  II. ***Occlusion Handling:*** Objects partially hidden in images pose challenges for accurate identification.
  III. ***Generalization:*** Models trained on specific datasets may struggle in unfamiliar environments.

### 3. Existing System

Object detection and tracking have been extensively studied in computer vision, leveraging traditional computer vision techniques and deep learning-based approaches. Existing systems typically employ machine learning models, convolutional neural networks (CNNs), and advanced tracking algorithms to detect and follow objects in various environments.

*3.1 Traditional Methods*
Earlier object detection systems relied on handcrafted features and classical machine learning models such as:

  I. ***Histogram of Oriented Gradients (HOG) + Support Vector Machines (SVM):*** Used for detecting objects like pedestrians in images.
  II. ***Viola-Jones Algorithm:*** A cascade classifier primarily used for face detection.
  III. ***Optical Flow & Kalman Filters:*** Used for tracking objects by estimating their motion in video sequences.

While these approaches worked well for specific tasks, they struggled with dynamic backgrounds, occlusions, and varying lighting conditions.

*3.2 Deep Learning-Based Methods*

Recent advancements in deep learning have significantly improved object detection and tracking accuracy. Common approaches include:

  I. ***Convolutional Neural Networks (CNNs):*** Models like AlexNet and VGG extract deep hierarchical features for object detection.
  II. ***Region-Based CNNs (R-CNN, Fast R-CNN, and Faster R-CNN):*** Use selective search or region proposal networks to enhance detection accuracy.
  III. ***You Only Look Once (YOLO) & Single Shot MultiBox Detector (SSD):*** Real-time object detection models that balance speed and accuracy.
  IV. ***Transformers (DETR):*** Attention-based models that replace CNNs for object detection, offering better long-range dependencies.

*3.3 Bounding Box*

The bounding box is a rectangle that has been drawn on the image and encircles the object.

Each instance of each object in the image has a bounding box. Four numbers (the center x, center y, width, and height) are projected for the box. A measurement of the separation between the predicted bounding box and the actual ground truth can be used to train this. The jacquard distance, which calculates the intersection of the predicted truth box and the ground, serves as the unit of measurement.

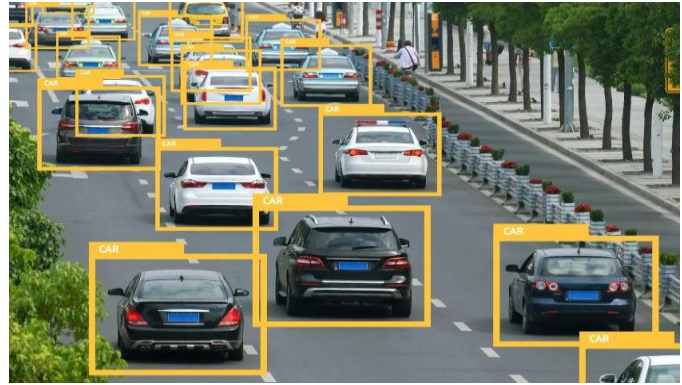

**Fig:3** Bounding Box-Based Vehicle Detection

The image illustrates an object detection system identifying multiple cars using yellow bounding boxes and labels. Each box highlights the position and classification of a detected car, enabling real-time tracking. Such systems are used in traffic monitoring, autonomous driving, and surveillance.

*3.4 Two-Stage Method*

The two-stage method is a widely used approach in object detection that separates the process into two distinct stages:
1. Region Proposal Stage:
   - The model first identifies potential object regions (Region of Interest, RoI).
   - This is done using algorithms like Selective Search or a Region Proposal Network (RPN) in Faster R-CNN.

2. Object Classification & Refinement Stage:
   - The proposed regions are passed to a deep learning classifier (e.g., CNN) to identify objects and refine bounding boxes.

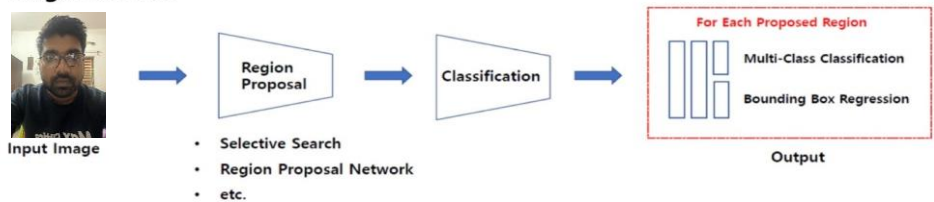
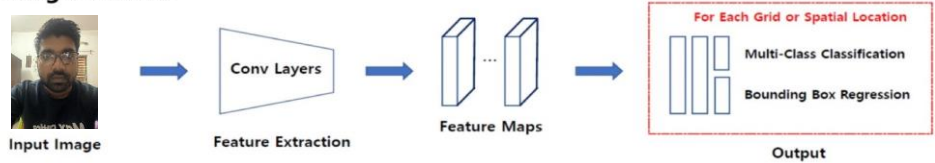

**Fig:4** Two-Stage Object Detector

## 4. Implementation and Results

We implemented an object detection model using YOLO, Faster R-CNN, SSD, etc. The datasets used for training and evaluation were COCO, Pascal VOC, and custom dataset.

### 4.1 Specifics of Implementation

Python 3 is used to implement the project. Deep network training was done with Tensor ow, and picture pre-processing was done with OpenCV.
The following are some technical details of the device used to train and test the model: Nvidia Titan Xp graphics card, Intel Core i7-7700 3.60 GHz CPU, and 32 GB of RAM.

### 4.2 Pre-processing

In order to speed up reading, the annotated data is presented in XML format, which is read and saved as a pickle file with the image. The image is also scaled down to a predetermined size.

## 5. Tensor Board Network

Tensor Board is a powerful visualization tool for Tensor Flow that helps monitor, debug, and optimize deep learning models, including object detection networks. When training an object detection model, Tensor Board provides insights into various aspects such as loss trends, learning rate schedules, model architecture, and detection results.

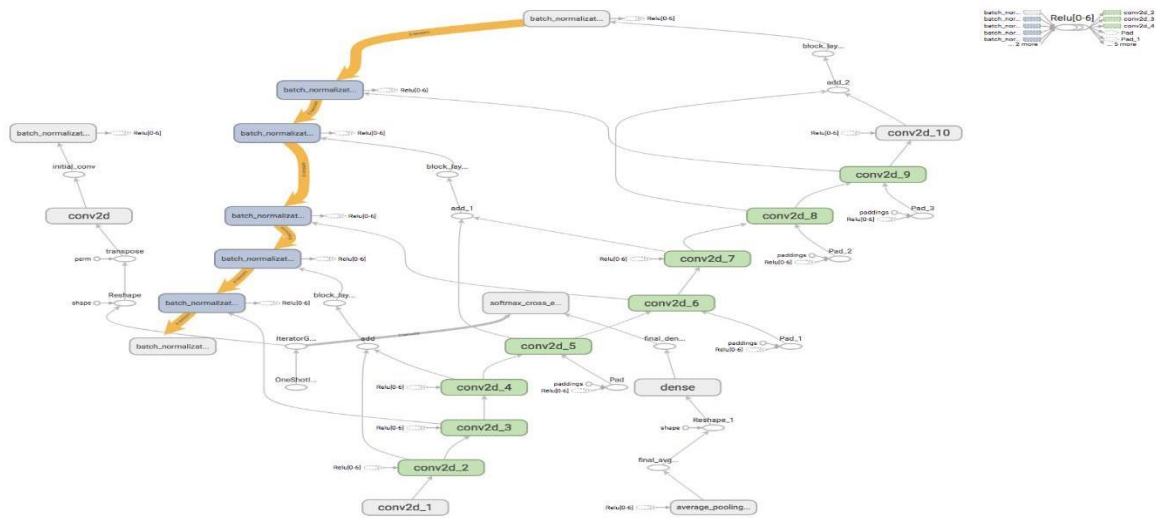

**Fig:5** Tensor Board Network

This image is a TensorBoard computation graph visualization of a deep learning model, specifically a convolutional neural network (CNN) used for object detection or classification.

### 5.1 Debugging & Optimizing Using Tensor Board Network

I. Identifying Computational Bottlenecks:
- Look for layers with high computation intensity (bold paths).
- Optimize by using techniques like depthwise separable convolutions or pruning.

II. Checking for Incorrect Layer Connections:

- Ensure input/output shapes match expected dimensions.
- Look for misplaced layers causing errors.
III. Analyzing Activation Functions & Batch Norm Layers:
- Check the placement of ReLU and Batch Normalization layers.
- Avoid issues like vanishing gradients in deep networks.

*5.2 Common Issues in Tensor Board Network Graphs*

| Issue | Cause | Solution |
|---|---|---|
| **Disconnected nodes** | Incorrect layer connections | Check input/output shapes |
| **Complex graphs** | Too many operations | Use tf.function for optimization |
| **Slow training** | Bottlenecks in large layers | Reduce layer complexity |
| **Vanishing gradients** | Deep networks without batch normalization | Use normalization and activation functions |

## 6. Tracking Approaches

*6.1 Traditional Tracking Methods*
Before deep learning, object tracking relied on hand-crafted features and mathematical models:

I. **Centroid Tracking**
- Tracks objects by calculating the centroid of the bounding box.
- Uses Euclidean distance to match objects between frames.
- Works well for simple scenarios but fails in occlusion and fast-moving objects.
II. **Kalman Filter-Based Tracking**
- A recursive algorithm that predicts an object's future position based on past states.
- Advantages: Works well under noise, and handles linear motion smoothly.
- Limitations: Assumes Gaussian noise and struggles with abrupt motion changes.
III. **Optical Flow Tracking**
- Estimates the motion of pixels between frames using techniques like Lucas-Kanade Optical Flow.
- *Advantages:* Does not require explicit detection in every frame.
- *Limitations:* Sensitive to illumination changes and background noise.
IV. **Mean-Shift and CAMShift Tracking**
- Tracks an object by iteratively shifting towards regions with similar color histograms.
- CAMShift (Continuously Adaptive Mean Shift) adapts to scale and rotation changes.
- Limitations: Fails under significant occlusion or background clutter.
V. **SORT (Simple Online and Realtime Tracker)**
- Combines Kalman filtering for motion prediction and the Hungarian algorithm for object association.
- Limitations: Cannot handle identity switches well.

## 7. Challenges in Object Tracking
Object tracking is a complex problem in computer vision, especially in real-world scenarios where factors such as occlusion, motion blur, and identity switches affect accuracy. Below are some of the most significant challenges faced in object tracking :

### 7.1 Occlusion (Partial or Full Object Blocking)
Occlusion occurs when an object being tracked is temporarily blocked by another object or disappears behind an obstacle. This makes it difficult for the tracker to maintain object identity.

### Types of Occlusion:
1. **Partial Occlusion:** Only part of the object is covered, making it difficult to extract features.
2. **Full Occlusion:** The object is completely hidden for a few frames, leading to identity loss.

### 7.1.1 Solution Approaches:

I. **Kalman Filters & Particle Filters:** Predict object positions even when occluded.
II. **Re-identification (ReID) Models:** Deep learning models match objects before and after occlusion.
III. **Transformer-Based Tracking:** Attention mechanisms help retain object identity.

### 7.2 Background Clutter & Similar Object Interference
Complex backgrounds make it difficult to distinguish objects from their surroundings.

### 7.2.1 Example
- Tracking a person wearing camouflage in a forest.
- Tracking a white car in snowy conditions.

### 7.2.2 Solution Approaches:
I. **Background Subtraction Techniques:** Helps remove static backgrounds.
II. **Instance Segmentation:** Distinguishes objects from the background.
III. **Feature Embedding Models:** Uses deep learning to differentiate similar objects.

### 7.3 Summary Table: Object Tracking Challenges & Solutions

| Challenge | Cause | Possible Solution |
|---|---|---|
| Occlusion | Object temporarily hidden | Re-identification models, Kalman Filters |
| Identity Switching | Similar objects confused | DeepSORT, Appearance-based tracking |
| Motion Blur | Fast-moving objects | Optical flow, Super-resolution models |
| Scale Variations | Objects change size | Adaptive tracking, Multi-scale CNNs |
| Background Clutter | Objects blend with the surroundings | Instance segmentation, Feature embedding |
| High Computational Cost | Deep models need high processing | Edge AI, Lightweight trackers |
| Long-Term Tracking Issues | Objects leave/re-enter the scene | Memory-augmented models, Re-detection |

## 8. Conclusion

With performance on par with current state-of-the-art systems, we have created an accurate and effective object-detecting system. The most recent computer vision and deep learning technologies are used in this project. Using labeling, I produced a unique data set, and the results were reliable. This can be utilized in real-time applications that call for object detection during the pipeline's pre-processing stage.
Training the system for video sequences for use in tracking applications is a crucial field. Frame-by-frame detection is less efficient than adding a time-coherent network, which provides seamless detection.

## 9. Future Work

Computer vision is still an evolving field and has not yet reached a level where it can be directly applied to real work. In a few years, computer vision and especially object detection will no longer be forward-looking, but ubiquitous. For now, you can think of object detection as a sub-branch of machine learning. Image AI provides more useful features if customization is required and production deployment if object detection is required. Some of the supported features:

I. *Adjusting minimum probability:* By default, objects detected with a percentage less than i50 are not displayed or reported. If you have high confidence, you can increase this value or decrease it if you need to detect all impossible objects to detect.
II. *Custom object detection:* You can use the Provided Custom Object class to report the discovery of one or more unique objects by specifying the discovery class.
III. *Detection Speed:* You can shorten the time it takes to detect an image by setting the detection speed for speed to "fast", "fastest" and "fastest".
IV. *Output Types:* You can specify that the Objects From image identification function returns an image from a form file or a Jumpy array.